\documentclass{article}

    \PassOptionsToPackage{numbers, compress}{natbib}
\usepackage{graphicx}

    \usepackage[preprint]{neurips_2025}

\usepackage[utf8]{inputenc} % allow utf-8 input
\usepackage[T1]{fontenc}    % use 8-bit T1 fonts
\usepackage{hyperref}       % hyperlinks
\usepackage{url}            % simple URL typesetting
\usepackage{booktabs}       % professional-quality tables
\usepackage{amsfonts}       % blackboard math symbols
\usepackage{nicefrac}       % compact symbols for 1/2, etc.
\usepackage{microtype}      % microtypography
\usepackage{xcolor}         % colors
\usepackage{multirow}

\usepackage{amsmath}
\usepackage{amssymb}
\usepackage{mathtools}
\usepackage{amsthm}
\usepackage{enumitem}
\theoremstyle{plain}

\theoremstyle{definition}

\theoremstyle{remark}

\usepackage{algorithm}
\usepackage{algpseudocode}
\usepackage{bm}

\usepackage{bbm}

\title{Sparse Autoencoders for Low-$N$ Protein Function Prediction and Design}

\author{%
Darin Tsui \\
Georgia Institute of Technology \\
darint@gatech.edu
\And
Kunal Talreja \\
Georgia Institute of Technology \\
ktalreja6@gatech.edu
\And
Amirali Aghazadeh \\
Georgia Institute of Technology \\
amiralia@gatech.edu
}

\begin{document}

\maketitle

\begin{abstract}
Predicting protein function from amino acid sequence remains a central challenge in data-scarce (low-$N$) regimes, limiting machine learning–guided protein design when only small amounts of assay-labeled sequence-function data are available. Protein language models (pLMs) have advanced the field by providing evolutionary-informed embeddings and sparse autoencoders (SAEs) have enabled decomposition of these embeddings into interpretable latent variables that capture structural and functional features. However, the effectiveness of SAEs for low-$N$ function prediction and protein design has not been systematically studied. Herein, we evaluate SAEs trained on fine-tuned ESM2 embeddings across diverse fitness extrapolation and protein engineering tasks. We show that SAEs, with as few as 24 sequences, consistently outperform or compete with their ESM2 baselines in fitness prediction, indicating that their sparse latent space encodes compact and biologically meaningful representations that generalize more effectively from limited data. Moreover, steering predictive latents exploits biological motifs in pLM representations, yielding top-fitness variants in 83\% of cases compared to designing with ESM2 alone.
\end{abstract}

\section{Introduction}

Machine learning (ML)–guided protein engineering seeks to predict and optimize protein function by leveraging evolutionary information and assay-labeled sequence data to model the underlying sequence–function landscape~\cite{yang2019machine, notin2024machine, ding2024machine}. In practice, however, ML models are often constrained by the scarcity of experimental data. Functional assays are costly and time-consuming, so only a small number of variants (low-$N$) can typically be characterized, creating a fundamental bottleneck for ML-guided design~\cite{biswas2021low, horvath2016screening, wu2019machine}.

Protein language models (pLMs), trained on large evolutionary sequence datasets, provide embeddings that achieve state-of-the-art performance in zero-shot function prediction~\cite{lin2023evolutionary, hayes2025simulating, Notin2022Tranception}. These embeddings are widely believed to capture amino acid interactions underlying protein function~\cite{tsui2024shapzero, tsui2024recovering, aghazadeh2021epistatic}, yet they remain difficult to interrogate. More recently, sparse autoencoders (SAEs) have emerged as a powerful interpretability framework, factorizing pLM embeddings into sparse, biologically meaningful latent variables. In high-$N$ regimes (e.g., $N > 800$ labeled sequences), these latents have been shown to align with structural and functional motifs~\cite{simon2024interplm, adams2025mechanistic, walton2025golf, gujral2025sparse} and can be steered to design sequences with targeted functional properties~\cite{parsan2025towards, garcia2025interpreting, corominas2025sparse}. Despite these advances, the function prediction and steering performance of SAEs in realistic data-scarce (low-$N$) settings has not been systematically evaluated. \emph{We hypothesize that the sparse latent space of SAEs, originally introduced as a strategy to enhance interpretability, also encodes compressed and regularized representations that enable accurate fitness prediction and effective protein design from limited data.} To test this, we reposition SAEs from proof-of-concept interpretability tools to actionable predictors and design engines, evaluating their performance on downstream protein engineering tasks under low-$N$ conditions. Specifically, we assess their utility across diverse fitness extrapolation challenges that reflect real-world design constraints, and we further examine their ability to design high-functioning variants through latent steering. Our main contributions are as follows:
\begin{itemize}
    \item We train SAEs on fine-tuned ESM2 embeddings across five proteins with diverse functions.  
    \item We show that SAEs, with as few as 24 sequences, outperform their ESM2 baselines in 58\% of fitness extrapolation tasks, while maintaining comparable performance in the remainder.  
    \item We demonstrate that steering SAEs along their most predictive latents produces a diverse pool of highly functional variants, including the top fitness variants in 83\% of cases, compared to designing with ESM2 alone.  
    \item We analyze the best-performing steered variants in green fluorescent protein (GFP) and the IgG-binding domain of protein G (GB1), uncovering biologically meaningful motifs that SAEs exploit for steering. All codes and data are available on our GitHub repository \url{https://github.com/amirgroup-codes/LowNSAE}.
\end{itemize}

\begin{figure*}[t!]
\vspace{-0cm}
\centering
\includegraphics[width=1\textwidth]{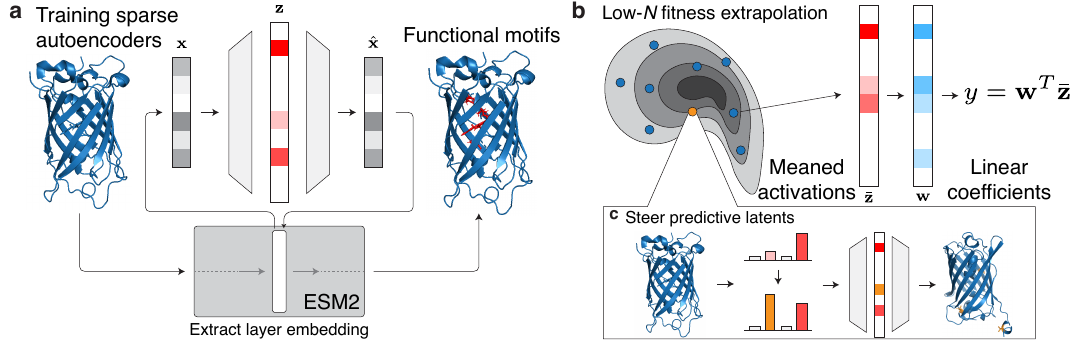}
\vspace{-0.3cm}
\caption{{\bf Overview of downstream low-$N$ tasks for SAEs.} {\bf a}, We train SAEs on the layer embeddings of ESM2. By projecting the model embedding $\mathbf{x}$ to the latent representation $\mathbf{z}$, and reconstructing the model embedding as $\hat{\mathbf{x}}$, the activations in $\mathbf{z}$ correspond to specific biological motifs. {\bf b}, In low-$N$ fitness extrapolation, a linear probe is trained on top of the SAE's latent space to predict protein fitness from $N$ many training sequences. {\bf c}, Using the learned linear probe weights, we steer predictive latents to design highly-functional variants. 
} 
% \vspace{-0.4cm} 
\label{fig:overview}
\end{figure*}
\section{Sparse Autoencoders (SAE)}

SAEs are autoencoders designed to learn meaningful representations of model embeddings in their latent space (Fig.~\ref{fig:overview}a). We use SAEs with TopK activation~\cite{gao2025scaling} to enforce sparsity in
the latent space. Given the model embedding $\mathbf{x} \in \mathbb{R}^{d_{\text{model}} \times L}$, where $L$ is the sequence length and $d_{\text{model}}$ is the embedding dimension, the encoder maps $\mathbf{x}$ to the SAE latent representation $\mathbf{z} \in \mathbb{R}^{d_{\text{SAE}} \times L}$ via:
\begin{equation}
    \label{eq:encoder}
    \mathbf{z} = \text{TopK}(\mathbf{W_{\text{enc}}}(\mathbf{x} - \mathbf{b_{\text{pre}}})),
\end{equation}
where $\mathbf{W_{\text{enc}}} \in \mathbb{R}^{d_{\text{SAE}} \times d_{\text{model}}}$ are the encoder weights and $ \mathbf{b_{\text{pre}}} \in \mathbb{R}^{d_{\text{model}} \times L}$ is a bias term. The TopK function is applied column-wise to the resulting matrix, keeping only the $k$ largest activations for each of the $L$ sequence positions and setting all other values to zero. The decoder then reconstructs the input $\mathbf{x}$ from $\mathbf{z}$ as: 
\begin{equation}
    \label{eq:decoder}
    \hat{\mathbf{x}} = \mathbf{W_{\text{dec}}} \mathbf{z} + \mathbf{b_{\text{pre}}}, 
\end{equation}
where $\mathbf{W_{\text{dec}}} \in \mathbb{R}^{d_{\text{model}} \times d_{\text{SAE}}}$ are the decoder weights. As illustrated in Fig.~\ref{fig:overview}a, where $L=1$ for simplicity, the activations in $\mathbf{z}$ have been shown to correspond to biological motifs~\cite{simon2024interplm, adams2025mechanistic}. 

During training, SAEs minimize both mean squared error and an auxiliary loss. The mean squared error between the original embedding $\mathbf{x}$ and its reconstruction $\hat{\mathbf{x}}$ is defined as $\mathcal{L}_{\text{MSE}} = \| \mathbf{x} - \hat{\mathbf{x}} \|_2^2$. To reduce the number of dead latents, defined as latents that never activate~\cite{gao2025scaling}, an auxiliary loss is included. Given the original reconstruction loss $\mathbf{e} = \mathbf{x} - \hat{\mathbf{x}}, $ the auxiliary loss is defined as $\mathcal{L}_{\text{aux}} = \| \mathbf{e} - \hat{\mathbf{e}} \|_2^2,$ where $\hat{\mathbf{e}}$ is found by multiplying the decoder matrix by the top-$k_{\text{aux}}$ latents in $\mathbf{z}$, where $k_{\text{aux}}$ is a hyperparameter. The total SAE training objective, $\mathcal{L}_{\text{SAE}}$, is a weighted sum of these two losses:
\begin{equation*}
\mathcal{L}_{\text{SAE}} = \mathcal{L}_{\text{MSE}} + \alpha \mathcal{L}_{\text{aux}},
\end{equation*}
where $\alpha$ is also a hyperparameter. This joint objective enables SAEs to not only reconstruct the original model embeddings faithfully, but also maximize the number of biologically interpretable latents.
% This joint objective encourages SAEs to not only reconstruct the original embeddings faithfully, but also to maximize the utility of their latent space by reducing the number of inactive dimensions. As a result, the learned representations are both compact and biologically interpretable, with a sparse set of latents carrying the most predictive signal.
% {\color{purple} and ... ??: you need a few sentences about the loss here. the section is ending too abruptly.}

\section{SAEs for Low-$N$ Fitness Extrapolation}

In this section, we first detail the datasets used and how we trained our SAEs. Then, we rigorously evaluate the ability of SAEs to generalize to unseen variants under various low-$N$ regimes (Fig.~\ref{fig:overview}b). To capture the challenges faced in real-world design settings, we define five distinct fitness extrapolation tasks that stress different aspects of the sequence–function landscape: random, position, mutation, regime, and score extrapolation. 

\subsection{Datasets and SAE Training Details}
\label{sec:training}

\textbf{Datasets.} We evaluated our SAEs on six deep mutational scanning (DMS) assays from ProteinGym~\cite{notin2023proteingym}, spanning five distinct proteins (Table~\ref{tab:dms}). These proteins were selected to ensure robust evaluation across a variety of functions. Additionally, these DMS assays also contain multipoint mutations, which are crucial for our fitness extrapolation tasks (see Section~\ref{sec:expt_setup_fitness}).

\begin{table*}[htbp]
\vspace{-0.2cm}
\caption{Summary of DMS assays used.}
\vspace{0.2cm}
\label{tab:dms}
\centering
\resizebox{\textwidth}{!}{%
\begin{tabular}{lcccccr}
\toprule
\textbf{DMS} & \textbf{Description} & \textbf{Function Tested} & \textbf{Variants} & \textbf{MSA Sequences} \\
\midrule
GFP\_AEQVI\_Sarkisyan~\cite{sarkisyan2016local}   & Green fluorescent protein & Fluorescence & 51,714 & 396 \\
SPG1\_STRSG\_Olson~\cite{olson2014comprehensive}  & IgG-binding domain of protein G  & Binding   & 536,962 & 44 \\
SPG1\_STRSG\_Wu~\cite{wu2016adaptation}  & IgG-binding domain of protein G  & Binding   & 149,360 & 3,109 \\
DLG4\_HUMAN\_Faure~\cite{faure2022mapping} & Third PDZ domain of PSD95  & Yeast growth       & 6,976 & 25,338 \\
GRB2\_HUMAN\_Faure~\cite{faure2022mapping} & C-terminal SH3 domain of GRB2  & Yeast growth       & 63,366 & 33,228 \\
F7YBW8\_MESOW\_Ding~\cite{ding2024protein}  & Antitoxin ParD3           & Growth enrichment  & 7,922 & 38,613 \\
\bottomrule
\end{tabular}%
}
\end{table*}

\textbf{Training.} For each DMS assay, we trained a unique SAE on a fine-tuned ESM2-650M model~\cite{lin2023evolutionary}. Each model was fine-tuned on multiple sequence alignment (MSA) sequences from Table~\ref{tab:dms} using  LoRA adapters, where the MSA sequences were obtained from ProteinGym. Embeddings $\mathbf{x}$ to train the SAE were then obtained by passing the MSA sequences through the fine-tuned model. Following~\cite{adams2025mechanistic}, we chose to extract embeddings from layer 24, and set $d_{\text{SAE}} = 4096, $ $k = 128,$ $\alpha = 1/32,$ and $k_{\text{aux}} = 256$. For more details on training, see Appendices~\ref{appendix:finetuning} and~\ref{appendix:saes}.

\subsection{Experimental Setup}
\label{sec:expt_setup_fitness}

\begin{figure*}[t!]
\vspace{-0cm}
\centering
\includegraphics[width=1\textwidth]{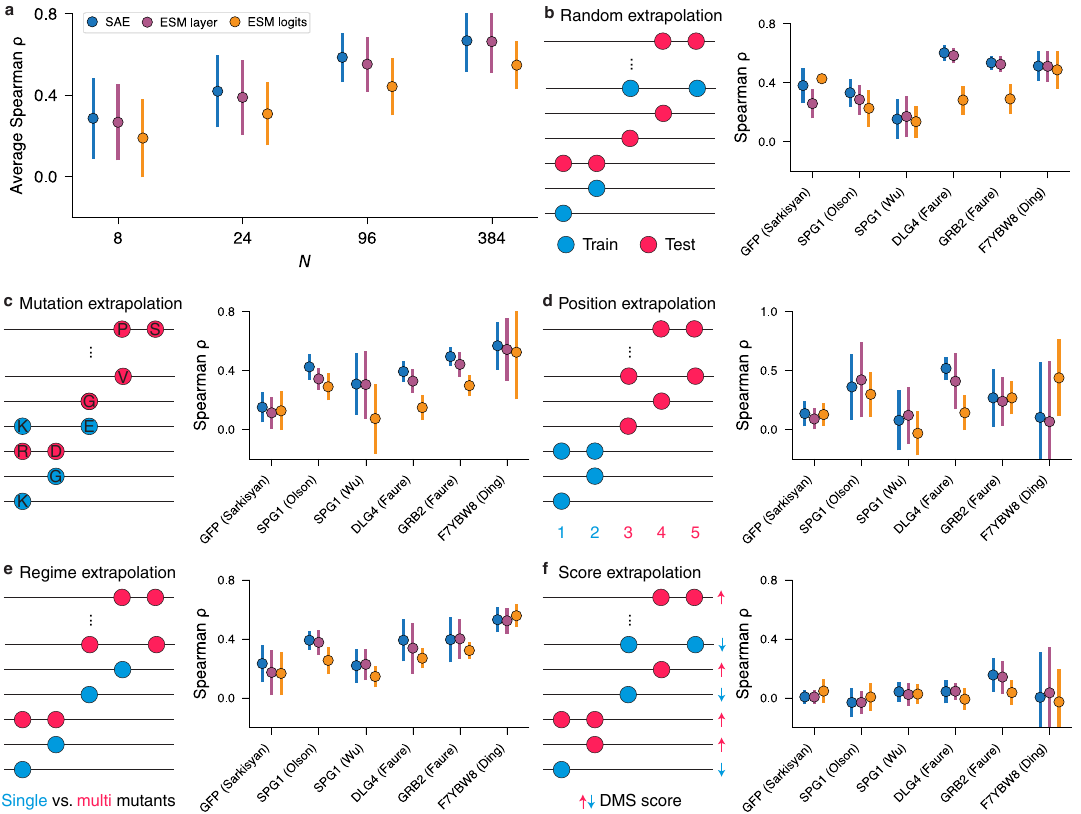}
\vspace{-0.3cm}
\caption{{\bf Comparative performance of low-$N$ fitness extrapolation in SAEs versus ESM2.} {\bf a}, Average correlations of SAE, ESM layer, and ESM logits across all low-$N$ regimes on random extrapolation. Fitness extrapolation correlations over each DMS assay using $N=24$ sequences across {\bf b}, random, {\bf c}, mutation, {\bf d}, position, {\bf e}, regime, and {\bf f}, score extrapolations. Error bars represent the standard deviation across nine independent runs with different random seeds.}
% \vspace{-0.4cm} 
\label{fig:extrapolation}
\end{figure*}

\textbf{Low-$N$ Regimes.} To evaluate the performance of our SAEs and ESM2 in the low-$N$ regime, we first created four distinct $N$ sizes to train a supervised model on top of the SAE latent space and ESM2 embeddings, respectively, to predict fitness: $N \in [8, 24, 96, 384]$. These sizes correspond to standard plate-well sizes used in protein engineering experiments~\cite{biswas2021low}.

\textbf{Fitness Extrapolation Tasks.} For each of our DMS assays, we designed five different fitness extrapolation tasks based on ref.~\cite {gelman2024biophysics} to test the ability of our SAE and ESM2 to generalize to unseen variants (see Fig.~\ref{fig:extrapolation}):
\begin{enumerate}
    \item \textbf{Random extrapolation}: We randomly sampled $N$ sequences from the DMS for training and validation, with 10\% of the DMS held out as a test set (Fig.~\ref{fig:extrapolation}b).
    \item \textbf{Mutation extrapolation}: We randomly designated 80\% of all possible mutations as training mutations (Fig.~\ref{fig:extrapolation}c). We sampled $N$ sequences for training that only had training mutations. The other 20\% of mutations were held out as a test set. 
    \item \textbf{Position extrapolation}: We randomly designated 80\% of amino acid positions as training positions (Fig.~\ref{fig:extrapolation}d). We then sampled $N$ sequences for training that had mutations exclusively at the training positions. The other 20\% of positions were held out as a test set. 
    \item \textbf{Regime extrapolation}: For DMS assays containing only single and double mutations, we trained on $N$ single mutations and tested on all double mutations. For DMS assays with more than two mutations, we trained on $N$ sequences drawn from single and double mutations, and tested on all sequences with more than two mutations (Fig.~\ref{fig:extrapolation}e). 
    \item \textbf{Score extrapolation}: We trained on $N$ sequences with a fitness score lower than the wildtype and tested on all sequences with a fitness score higher than the wildtype (Fig.~\ref{fig:extrapolation}f).
\end{enumerate}

\textbf{Linear Probes.} For each of these extrapolations, we trained a linear probe with Ridge regression on top of the SAE latent space. To benchmark against the performance of ESM2 without help from an SAE, we also trained linear probes on the ESM2 layer 24 embedding and ESM2 logits. For brevity, we refer to these methods as 1) SAE, 2) ESM layer, and 3) ESM logits. Following~\cite{adams2025mechanistic}, we mean-pool the input to the linear probe over the respective embedding dimension. Formally, in SAEs, we denote $\bar{\mathbf{z}} \in \mathbb{R}^{d_{\text{SAE}}}$ to be the meaned activations of the latent space and $\mathbf{w} \in \mathbb{R}^{d_{\text{SAE}}}$ to be weights of the linear probe. The linear probe then computes the fitness score $y$ via: $y= \mathbf{w}^T \bar{\mathbf{z}}$ (Fig.~\ref{fig:overview}b). For all tasks, we set aside a portion of the training sequences to be used as validation. Further details are provided in Appendix~\ref{appendix:extrapolation}.

To ensure the robustness of our results, we ran a total of nine trials for each extrapolation. For random, position, and mutation extrapolations, we used three different random seeds to create the test set. For each of these test sets, we then randomly sampled $N$ training sequences three times. For the regime and score extrapolations, where the test set is deterministic, we randomly sampled $N$ training sequences nine times.

\subsection{SAEs Achieve Improved Generalization to Unseen Variants Compared to ESM2}
\label{sec:results_fitness}

Fig.~\ref{fig:extrapolation}a shows the average Spearman correlation of SAE, ESM layer, and ESM logits under random extrapolation across all low-$N$ regimes, while Table~\ref{tab:learning_curve} breaks down results by DMS assay. SAEs achieve higher correlations than their ESM2 counterparts in 67\% of random extrapolation experiments, and across all low-$N$ regimes and fitness extrapolation tasks (Appendix~\ref{appendix:expt_results}), they outperform in 58\% of cases. These results suggest that SAE latents capture more biologically meaningful patterns and enable more reliable generalization to unseen variants. Among the different extrapolation settings, position, regime, and score extrapolation emerge as the most challenging, since they require the model to capture structural context and nonlinear interactions underlying protein function. Notably, SAEs outperform their ESM2 counterparts in 69\% of position and regime extrapolation tasks, suggesting that their sparse latent space encodes fundamental biological constraints. We additionally notice that SAEs are not able to generalize to unseen variants when ESM2 does a poor job, such as in score extrapolation. This is not surprising, as intuitively, SAEs are trained to reorganize the information encoded by ESM2 into a more compact and disentangled representation. Thus, when the underlying pLM provides a limited predictive signal, the bottleneck in performance lies in the pLM rather than in the SAE.

\begin{table*}[h]
\vspace{-0.2cm}
\caption{Average Spearman $\rho$ across all low-$N$ regimes under random extrapolation across each DMS assay. A full summary of results for other fitness extrapolation tasks is located in Appendix~\ref{appendix:expt_results}.}
\vspace{0.2cm}
\label{tab:learning_curve}
\centering
\resizebox{1 \textwidth}{!}{%
\begin{tabular}{lrcccc}
\toprule
\textbf{Method} & \textbf{DMS}  & $\boldsymbol{N}\mathbf{=8} \,\uparrow$  & $\boldsymbol{N}\mathbf{=24} \,\uparrow$  & $\boldsymbol{N}\mathbf{=96} \,\uparrow$  & $\boldsymbol{N}\mathbf{=384} \,\uparrow$ \\
\midrule
\multirow{6}{*}{SAE}  & GFP\_AEQVI\_Sarkisyan & 0.26 $\pm$ 0.15 & 0.38 $\pm$ 0.12 & \textbf{0.56 $\pm$ 0.05} & \textbf{0.67 $\pm$ 0.01} \\
  & SPG1\_STRSG\_Olson & 0.16 $\pm$ 0.17 & \textbf{0.33 $\pm$ 0.09} & \textbf{0.67 $\pm$ 0.03} & \textbf{0.82 $\pm$ 0.01} \\
  & SPG1\_STRSG\_Wu & 0.12 $\pm$ 0.16 & 0.15 $\pm$ 0.14 & \textbf{0.34 $\pm$ 0.04} & 0.35 $\pm$ 0.03 \\
  & DLG4\_HUMAN\_Faure & \textbf{0.45 $\pm$ 0.16} & \textbf{0.60 $\pm$ 0.05} & \textbf{0.67 $\pm$ 0.03} & \textbf{0.76 $\pm$ 0.02} \\
  & GRB2\_HUMAN\_Faure & \textbf{0.31 $\pm$ 0.15} & \textbf{0.53 $\pm$ 0.05} & \textbf{0.63 $\pm$ 0.02} & \textbf{0.73 $\pm$ 0.01} \\
  & F7YBW8\_MESOW\_Ding & \textbf{0.42 $\pm$ 0.19} & \textbf{0.51 $\pm$ 0.10} & 0.64 $\pm$ 0.02 & 0.68 $\pm$ 0.02 \\
\cmidrule(lr){1-6}
\multirow{6}{*}{ESM layer}  & GFP\_AEQVI\_Sarkisyan & 0.21 $\pm$ 0.13 & 0.26 $\pm$ 0.10 & 0.46 $\pm$ 0.07 & 0.61 $\pm$ 0.01 \\
  & SPG1\_STRSG\_Olson & \textbf{0.17 $\pm$ 0.22} & 0.28 $\pm$ 0.10 & 0.64 $\pm$ 0.02 & 0.81 $\pm$ 0.01 \\
  & SPG1\_STRSG\_Wu & \textbf{0.13 $\pm$ 0.15} & \textbf{0.17 $\pm$ 0.14} & 0.31 $\pm$ 0.05 & \textbf{0.36 $\pm$ 0.07} \\
  & DLG4\_HUMAN\_Faure & 0.40 $\pm$ 0.16 & 0.58 $\pm$ 0.05 & 0.66 $\pm$ 0.05 & \textbf{0.76 $\pm$ 0.02} \\
  & GRB2\_HUMAN\_Faure & 0.28 $\pm$ 0.14 & 0.52 $\pm$ 0.05 & 0.60 $\pm$ 0.03 & 0.73 $\pm$ 0.02 \\
  & F7YBW8\_MESOW\_Ding & 0.40 $\pm$ 0.15 & 0.51 $\pm$ 0.11 & \textbf{0.65 $\pm$ 0.02} & \textbf{0.70 $\pm$ 0.01} \\
\cmidrule(lr){1-6}
\multirow{6}{*}{ESM logits}  & GFP\_AEQVI\_Sarkisyan & \textbf{0.31 $\pm$ 0.12} & \textbf{0.43 $\pm$ 0.03} & 0.49 $\pm$ 0.05 & 0.57 $\pm$ 0.03 \\
  & SPG1\_STRSG\_Olson & 0.12 $\pm$ 0.13 & 0.23 $\pm$ 0.13 & 0.42 $\pm$ 0.02 & 0.55 $\pm$ 0.01 \\
  & SPG1\_STRSG\_Wu & 0.01 $\pm$ 0.13 & 0.14 $\pm$ 0.11 & 0.21 $\pm$ 0.08 & 0.30 $\pm$ 0.04 \\
  & DLG4\_HUMAN\_Faure & 0.17 $\pm$ 0.17 & 0.28 $\pm$ 0.10 & 0.47 $\pm$ 0.05 & 0.60 $\pm$ 0.04 \\
  & GRB2\_HUMAN\_Faure & 0.14 $\pm$ 0.10 & 0.29 $\pm$ 0.10 & 0.41 $\pm$ 0.07 & 0.59 $\pm$ 0.02 \\
  & F7YBW8\_MESOW\_Ding & 0.38 $\pm$ 0.25 & 0.49 $\pm$ 0.13 & 0.65 $\pm$ 0.03 & 0.67 $\pm$ 0.02 \\
%\cmidrule(lr){1-6}
\bottomrule
\end{tabular}%
}
\end{table*}

Fig.~\ref{fig:extrapolation}b-f further illustrates the performance of SAE, ESM layer, and ESM logits across all extrapolation tasks with $N=24$ sequences. We designated this as the smallest low-$N$ regime for reliable extrapolation, with the SAE achieving an average correlation of 0.42. Across nearly all tasks, SAEs either match or outperform both ESM layers and ESM logits, highlighting their robustness and effectiveness in diverse low-$N$ extrapolation settings. For additional results, see Appendix~\ref{appendix:expt_results}.

\section{SAEs for Low-$N$ Protein Engineering}
\label{sec:results_protein_engineering}

After demonstrating that SAEs are able to generalize to unseen variants, we then looked to assess their performance in generating high-functioning proteins (Fig.~\ref{fig:overview}c). To explicitly optimize for function, we implemented a modified version of feature steering~\cite{templeton2024scaling}, which leverages the predictive scores from the linear probes. For all experiments, we used the linear probes trained on $N=24$ sequences. 

\subsection{Experimental Setup} 
\label{sec:expt_setup_protein_engineering}

For feature steering, we first identified the most predictive latent features by examining the largest-magnitude weights of the linear probe. For each of these high-impact latents, we increased its activation by a hyperparameter multiplier. The updated latent representation was then passed through the SAE decoder and fed into ESM2 to design a new sequence. We optimized the multiplier by selecting the value that yielded the highest predicted fitness score from the linear probe. Similar to fitness extrapolation, to benchmark against the performance of ESM2, we also designed sequences using the linear probes trained on the ESM layer and ESM logits via simulated annealing, following the procedure detailed in~\cite{gelman2024biophysics}. Additionally, we included a random baseline by generating sequences with a random number of mutations and amino acid substitutions. Further details on our experimental setup are provided in Appendix~\ref{appendix:protein_engineering}.

We used a multi-layer perceptron (MLP) trained on the DMS assays to evaluate the fitness of our designed variants (see Appendix~\ref{appendix:mlp}). To constrain our search space and ensure the MLP's predictions are a good proxy for experimental fitness, we limited all designed variants to a maximum of five mutations away from the wildtype. A notable exception to this setup is the SPG1\_STRSG\_Wu DMS: this assay provides ground-truth fitness values for all possible combinatorial variants over four positions. Therefore, we directly used the fitness values from SPG1\_STRSG\_Wu to evaluate our designed variants and limited our maximum number of mutations to four. A total of 50 variants were designed per DMS assay.

\subsection{SAEs Design High-functional Variants and Capture Biological Motifs}
\label{sec:results_fitness}

Table~\ref{tab:protein_engineering} shows the performance of all methods in generating highly-functional variants. Across all metrics and DMS assays, SAEs outperform their ESM2 counterparts in 88\% of cases. More specifically, our SAE steering approach designed the top fitness variants in five out of the six DMS assays. Additionally, SAE steering designed the highest top 10\% fitness variants across all DMS assays and the highest top 20\% variants in five out of six DMS assays. This suggests that SAE steering is not only capable of discovering the single top-performing variant, but also is capable of generating a diverse pool of highly functional variants.

\begin{table*}[t]
\caption{Protein engineering results using $N=24$ training sequences. All variants were constrained to a maximum of five mutations away from the wild type.}
\vspace{0.2cm}
\label{tab:protein_engineering}
\centering
\resizebox{1 \textwidth}{!}{%
\begin{tabular}{lrccccc}
\toprule
\textbf{Method} & \textbf{DMS}
  & \textbf{Mean fitness $\uparrow$} & \textbf{Max fitness $\uparrow$}
  & \textbf{Top 10\% fitness $\uparrow$} & \textbf{Top 20\% fitness $\uparrow$} \\
\midrule
\multirow{1}{*}{SAE}
  & GFP\_AEQVI\_Sarkisyan  & \textbf{3.49} $\pm$ 0.44 & \textbf{3.87} & \textbf{3.75} $\pm$ 0.08 & 3.71 $\pm$ 0.07 \\
  & SPG1\_STRSG\_Olson  & \textbf{2.75} $\pm$ 1.29 & \textbf{4.53} & \textbf{4.47} $\pm$ 0.04 & \textbf{4.29} $\pm$ 0.24 \\
  & SPG1\_STRSG\_Wu  & \textbf{0.67} $\pm$ 0.94 & \textbf{3.89} & \textbf{2.70} $\pm$ 0.79 & \textbf{2.18} $\pm$ 0.76 \\
  & DLG4\_HUMAN\_Faure  & \textbf{0.39} $\pm$ 0.22 & \textbf{0.68} & \textbf{0.66} $\pm$ 0.02 & \textbf{0.62} $\pm$ 0.05 \\
  & GRB2\_HUMAN\_Faure  & \textbf{-0.10} $\pm$ 0.48 & \textbf{0.67} & \textbf{0.59} $\pm$ 0.07 & \textbf{0.49} $\pm$ 0.12 \\
  & F7YBW8\_MESOW\_Ding  & 0.81 $\pm$ 0.33 & 1.16 & \textbf{1.15} $\pm$ 0.01 & \textbf{1.13} $\pm$ 0.03 \\
\cmidrule(lr){1-6}
\multirow{1}{*}{ESM layer}
  & GFP\_AEQVI\_Sarkisyan  & 3.29 $\pm$ 0.66 & 3.72 & 3.71 $\pm$ 0.01 & 3.70 $\pm$ 0.01 \\
  & SPG1\_STRSG\_Olson  & 0.29 $\pm$ 1.95 & 3.19 & 2.74 $\pm$ 0.35 & 2.44 $\pm$ 0.39 \\
  & SPG1\_STRSG\_Wu  & 0.08 $\pm$ 0.30 & 1.69 & 0.81 $\pm$ 0.63 & 0.41 $\pm$ 0.60 \\
  & DLG4\_HUMAN\_Faure  & -0.10 $\pm$ 0.41 & 0.63 & 0.45 $\pm$ 0.14 & 0.36 $\pm$ 0.13 \\
  & GRB2\_HUMAN\_Faure  & -0.40 $\pm$ 0.39 & 0.30 & 0.24 $\pm$ 0.05 & 0.17 $\pm$ 0.10 \\
  & F7YBW8\_MESOW\_Ding  & \textbf{1.06} $\pm$ 0.10 & \textbf{1.16} & 1.15 $\pm$ 0.02 & 1.12 $\pm$ 0.03 \\
\cmidrule(lr){1-6}
\multirow{1}{*}{ESM logits}
  & GFP\_AEQVI\_Sarkisyan  & 3.13 $\pm$ 0.86 & 3.76 & 3.73 $\pm$ 0.02 & \textbf{3.72} $\pm$ 0.02 \\
  & SPG1\_STRSG\_Olson  & -1.11 $\pm$ 2.21 & 2.27 & 2.05 $\pm$ 0.42 & 1.56 $\pm$ 0.60 \\
  & SPG1\_STRSG\_Wu  & 0.15 $\pm$ 0.37 & 1.69 & 1.13 $\pm$ 0.33 & 0.76 $\pm$ 0.50 \\
  & DLG4\_HUMAN\_Faure  & -0.15 $\pm$ 0.38 & 0.53 & 0.36 $\pm$ 0.14 & 0.27 $\pm$ 0.13 \\
  & GRB2\_HUMAN\_Faure  & -0.26 $\pm$ 0.44 & 0.58 & 0.44 $\pm$ 0.09 & 0.32 $\pm$ 0.14 \\
  & F7YBW8\_MESOW\_Ding  & 1.05 $\pm$ 0.06 & 1.12 & 1.11 $\pm$ 0.01 & 1.11 $\pm$ 0.01 \\
\cmidrule(lr){1-6}
\multirow{1}{*}{Random}
  & GFP\_AEQVI\_Sarkisyan  & 3.36 $\pm$ 0.70 & 3.75 & 3.72 $\pm$ 0.02 & 3.70 $\pm$ 0.02 \\
  & SPG1\_STRSG\_Olson  & -1.25 $\pm$ 2.63 & 3.05 & 2.40 $\pm$ 0.49 & 1.99 $\pm$ 0.55 \\
  & SPG1\_STRSG\_Wu  & 0.33 $\pm$ 0.76 & 3.61 & 2.27 $\pm$ 0.82 & 1.36 $\pm$ 1.11 \\
  & DLG4\_HUMAN\_Faure  & -0.34 $\pm$ 0.43 & 0.35 & 0.32 $\pm$ 0.04 & 0.24 $\pm$ 0.10 \\
  & GRB2\_HUMAN\_Faure  & -0.96 $\pm$ 0.38 & 0.16 & -0.13 $\pm$ 0.18 & -0.30 $\pm$ 0.22 \\
  & F7YBW8\_MESOW\_Ding  & 0.66 $\pm$ 0.37 & 1.16 & 1.11 $\pm$ 0.03 & 1.06 $\pm$ 0.06 \\
\cmidrule(lr){1-6}
\bottomrule
\end{tabular}%
}
\end{table*}

To better understand why feature steering designs high-functional variants, we performed a qualitative analysis on the top-performing variants for the green fluorescent protein (GFP) and the IgG-binding domain of protein G (GB1). We identified the ten latent dimensions most strongly associated with changes in fitness. We also analyzed any shifts in their activation patterns between the wildtype and the designed variant. Finally, we projected these activated residues onto the variant's structure (generated via AlphaFold3~\cite{abramson2024accurate}) to identify amino acid concentrations and infer their biological relevance. Further details are provided in Appendix~\ref{appendix:feat-vis}.

\begin{figure*}[t!]
\vspace{-0cm}
\centering
\includegraphics[width=0.9\textwidth]{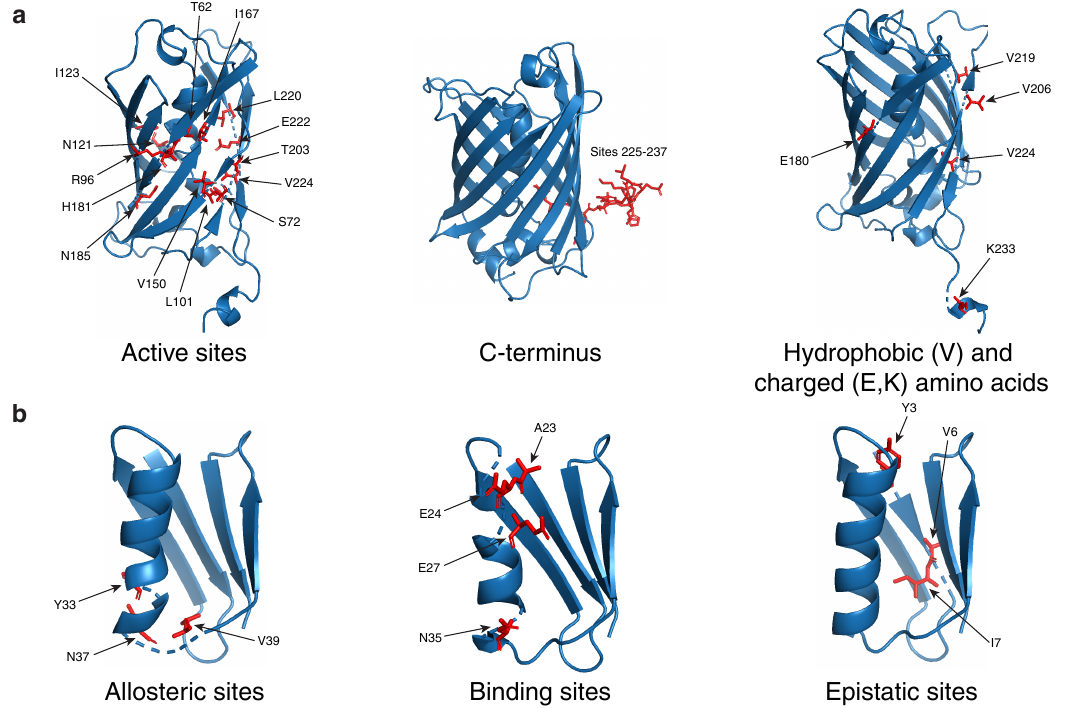}
\vspace{-0.1cm}
\caption{{\bf Analysis of the top-performing steered variants.} {\bf a}, Our analysis of the top-performing GFP variants revealed that steering activated latent features corresponding to key biological motifs, including active site amino acids, the C-terminus, and hydrophobic and charged amino acids. {\bf b}, GB1 variants activated latent features associated with allosteric, binding, and epistatic sites.}
% \vspace{-0.4cm} 
\label{fig:interpret}
\end{figure*}

Our analysis revealed that SAEs preferentially activated latent features associated with known biological motifs. For instance, in GFP, this includes latents activating on active site amino acids, which are crucial for fluorescence~\cite{weinstein2023designed}, and the C-terminus, a disordered region also known to affect fluorescence (Fig.~\ref{fig:interpret}a)~\cite{kim1998truncated}. We also found that steering favored latent features corresponding to hydrophobic and charged amino acids, which are essential for maintaining the protein's structural stability~\cite{ormo1996crystal}. Similarly, our analysis of the top-performing variants in GB1 highlighted key functional regions (Fig.~\ref{fig:interpret}b). The top latents were most active at sites that are allosteric~\cite{faure2022mapping}, which modulate protein function, or binding, which directly interact with the protein IgG~\cite{olson2014comprehensive}. Furthermore, latents activated on epistatic sites~\cite{olson2014comprehensive}, demonstrating the SAE's ability to design variants in the presence of complex, non-additive mutations. These findings collectively demonstrate that SAEs, even without explicit training, successfully learn and leverage fundamental biological principles to design new variants.

\section{Discussion}
\label{sec:discussion}

In this paper, we demonstrated that sparse autoencoders (SAEs) can serve as a powerful tool for low-$N$ tasks. We demonstrated that SAEs consistently outperform their ESM2 counterparts in a variety of low-$N$ fitness extrapolation tasks and are highly effective for generating novel, high-fitness protein variants. Our work expands the biologist's toolkit for resource-constrained applications and takes the first step toward extracting actionable biological knowledge from pLMs.

\begin{figure*}[t!]
\vspace{-0cm}
\centering
\includegraphics[width=1\textwidth]{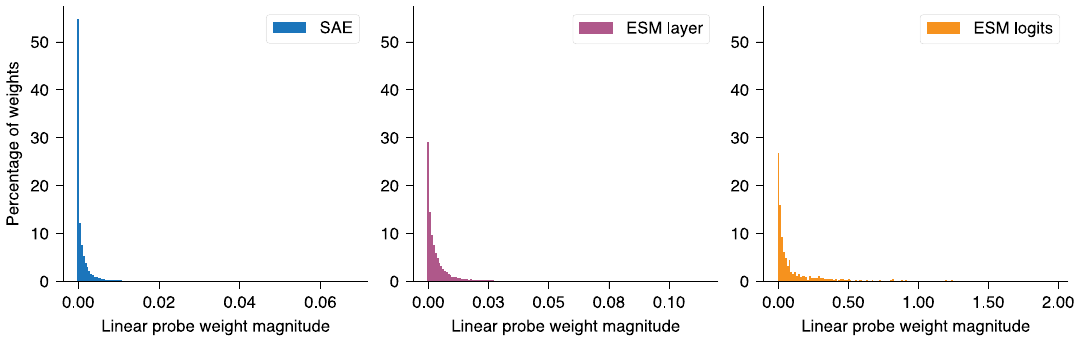}
\vspace{-0.3cm}
\caption{{\bf Sparsity in SAEs underlies improved low-$N$ performance.} 
Histogram of linear probe weight magnitudes for the SAE latent space, ESM layer embeddings, and ESM logits. Given the top 5\% of weights by magnitude, SAE weights explain $37 \pm 9\%$ of the variance, compared to $27 \pm 4\%$ in ESM layer weights and $25 \pm 12\%$ in ESM logits weights. See Appendix~\ref{appendix:weight-sparsity} for details on the visualization procedure.}
% \vspace{-0.4cm} 
\label{fig:histogram}
\end{figure*}

\textbf{Sparsity in SAEs.} Although SAEs introduce a larger dimensionality to the linear probes, they consistently outperform their ESM2 counterparts in low-$N$ fitness extrapolation and protein engineering tasks. At first glance, this appears counterintuitive: in low-$N$ regimes, simpler models with fewer parameters are typically less prone to overfitting. We attribute the superior performance of SAEs to their ability to compress biologically relevant information into a \emph{sparse} latent space (Fig.~\ref{fig:histogram}). To quantify this effect, we measure the variance explained by the magnitude of the top 5\% of probe weights. Under this definition, SAE weights explain $38 \pm 9\%$ of the variance, whereas ESM layer and ESM logits weights explain $28 \pm 3\%$ and $31 \pm 17\%$, respectively. These results suggest that SAEs compress information from ESM2 into a more compact and disentangled representation, where the biological relevant signal is concentrated in a select few latents. In the low-$N$ regime, this compression is particularly advantageous: sparser models are less prone to overfitting and thus generalize more effectively from limited experimental data. This also allows each high-impact latent to disentangle which amino acids contribute to fitness, enhancing the effectiveness of our steering approach.

\textbf{Low-$N$ Performance Variability.} Across fitness extrapolation tasks, we observe relatively high standard deviation in all methods, This is not surprising, given that the linear probes are trained in the low-$N$ regime, making the performance sensitive to which sequences are sampled. Despite this variability, we emphasize that SAEs outperform their ESM2 counterparts in 58\% of cases on average, indicating a consistent advantage. Moreover, SAEs tend to design more high-functioning variants, suggesting that their sparse latent space captures a more informative view of the functional landscape.

\textbf{Limitations and Future Work.} Our work analyzes the performance of SAEs across a wide range of proteins and molecular functions. However, our evaluation could be extended to other proteins with clinically relevant functions such as antibiotic resistance and viral replication, which may open new avenues for therapeutic design. We also observed that SAE performance is strongly influenced by the number of MSA sequences available for training. For example, proteins such as DLG4, GRB2, and F7YBW8, which consistently achieved high fitness extrapolation correlations, each had more than 20,000 MSA sequences. Future work should explore strategies for robust SAE training when MSAs are shallow or unavailable.

% Our protein engineering experiments relied on fitness models \emph{in silico}, which provide useful proxies but cannot fully substitute for experimental validation. Wet-lab assays remain necessary to confirm the functionality of designed variants. Encouragingly, in the SPG1_STRSG_Wu dataset—where ground-truth fitness values are available for all combinatorial variants—SAE steering produced the highest-fitness sequences among all methods. Still, successful design requires not only functional but also stable variants. Some proteins may exhibit high activity but low stability, underscoring the importance of integrating stability constraints. A promising direction is to couple SAE steering with physics-based tools such as Rosetta to jointly optimize for both function and stability.

To validate the designed variants, our protein engineering experiments rely on trained fitness models \emph{in silico}. While this is a good proxy for fitness, further validation through wet lab experiments is necessary to verify the function of designed variants. Nevertheless, our results on the SPG1\_STRSG\_Wu DMS assay, for which ground-truth fitness values are available for all combinatorial variants, demonstrate that our SAE steering approach still produces the highest-fitness variants compared to other methods. Additionally, we clarify that successful protein design requires not only highly-functional, but also highly-stable variants. Certain mutations that promote function may also destabilize the protein~\cite{shoichet1995relationship, meiering1992effect}, reinforcing the need for wet lab experiments to test designs. A promising future direction is to couple SAE steering with physics-based tools such as Rosetta~\cite{leman2020macromolecular} to jointly optimize for both function and stability.

Lastly, in our protein engineering experiments, we constrained variants to be a maximum of five mutations away from the wildtype. Beyond this radius, we found it difficult to design highly-functional sequences using just one predictive latent. Additionally, because predictive information is concentrated in a small number of latents, we restricted the amount of variants designed to 50 per DMS assay. Future work towards expanding the design space could involve steering multiple latents at once, which could enable both further mutational exploration and more diverse pools of functional variants. However, we note that in silico evaluation tools become less reliable the further away variants from the wildtype are, reinforcing the need for wet lab validation.

\bibliographystyle{unsrtnat} \bibliography{references}

\newpage
\appendix
\section{Additional Experimental Details}
\label{appendix:expt_setup}

\subsection{Fine-tuning ESM2}
\label{appendix:finetuning}

We fine-tuned the pre-trained ESM-2-650M model on the MSA of each DMS assay using LoRA (Low-Rank Adaptation) adapters to each layer. For each DMS assay, we loaded its corresponding MSA and masked 15\% of the amino acids in each sequence, consistent with ESM2's original masked language modeling objective. 

To prevent overfitting, we subsample the MSA by randomly selecting up to 1000 sequences to fine-tune on. The number of fine-tuning epochs was dynamically determined based on the number of sequences used. The epoch schedule was set as follows:
\begin{itemize}
    \item $<$ 100 sequences: 20 epochs
    \item 100-299 sequences: 10 epochs
    \item 300-499 sequences: 5 epochs
    \item 500-799 sequences: 4 epochs
    \item $\geq$ 800 sequences: 3 epochs
\end{itemize}
We fine-tuned the model using the AdamW optimizer with a learning rate of $10^{-4}$. We set the hyperparameters of LoRA to be the following: \texttt{r=8}, \texttt{lora\_alpha=16}, \texttt{lora\_dropout=0.05}, and \texttt{bias=None}.

\subsection{Training SAEs}
\label{appendix:saes}

We trained a unique SAE for each of the DMS assays. For each DMS assay, we first load in the respective fine-tuned ESM2 model (Appendix \ref{appendix:finetuning}). We adapted code from \url{https://github.com/etowahadams/interprot}~\cite{adams2025mechanistic} to train on embeddings from layer 24. We dynamically set the number of training epochs based on the number of MSA sequences in each assay. The epoch schedule was set as follows:
\begin{itemize}
    \item $<$ 500 sequences: 1000 epochs
    \item 500-999 sequences: 500 epochs
    \item 1000-4999 sequences: 100 epochs
    \item $\geq$ 5000 sequences: 10 epochs
\end{itemize}

\subsection{Fitness Extrapolation}
\label{appendix:extrapolation}
For all tasks except regime extrapolation, we set aside 10\% of the training sequences to use as a validation set and perform a grid search over regularization strengths. In regime extrapolation, we set aside 20\% as validation.  Since the dataset for F7YBW8\_MESOW\_Ding only has 166 single and double mutations, we modify the regime split to train on single, double, and triple mutations, and test on datapoints with more than three mutations. Since the DMS for SPG1\_STRSG\_Wu has only 4 sites with mutations, we also modify the position split to instead take 75\% of amino acid positions as training positions and 25\% as test.  

A full summary of results can be found in Appendix~\ref{appendix:expt_results}. In each fitness extrapolation table, we report the absolute value of the Spearman correlation plus the standard deviation across all nine trials. 

\subsection{Protein Engineering}
\label{appendix:protein_engineering}

In this section, we provide additional details on our protein engineering experimental setup. To ensure our trained MLPs can properly score the designed variants, we only design mutations at positions that are present in the DMS assays.

\textbf{Feature Steering.} Given the wildtype sequence, we first pass it through ESM2 to get the layer embeddings, and then pass it through the SAE encoder to get the latent representation $\mathbf{z}$. For the $i^\text{th}$ predictive latent, we multiply the $i^\text{th}$ row of $\mathbf{z}$ by a hyperparameter multiplier. The modified latent vector is then passed through the SAE decoder. The resulting vector is fed through the remaining layers of ESM2 to output the logits of the mutated sequence, $\mathbf{x}_\text{logits,mut} \in \mathbb{R}^{L \times V},$ where $L$ is the sequence length and $V$ is the vocabulary size.

We compare these new logits to the logits of the wild-type sequence, $\mathbf{x}_\text{logits,wt} \in \mathbb{R}^{L \times V}$. For each amino acid position, we calculate the cosine similarity between the respective logit vectors. We only accept a mutation at a given position if the cosine similarity is below 0.98, ensuring that we only mutate amino acids where ESM2 has made a meaningful change.
% To ensure our MLP can properly score the mutations, we additionally only accept mutations that are present in the deep mutational scanning (DMS) dataset.

To find the optimal multiplier, we perform a grid search over values from -3 to 3 with a step size of 0.2. For each multiplier, we use the linear probe to predict the fitness of the resulting sequence. We then select the top 50 unique sequences with the highest predicted fitness. If a sequence has been previously designed, we move to the next highest-scoring sequence to ensure we design 50 unique variants.

\textbf{Simulated Annealing.} We adapt the simulated annealing code from \url{https://github.com/gitter-lab/metl-pub/tree/main/sim-annealing}~\cite{gelman2024biophysics}. All parameters are left as default. We run simulated annealing over the linear probes trained on the ESM layer and ESM logits. The number of mutations per designed sequence was determined by sampling from the Poisson distribution $\text{Pois}(2) + 1$, ensuring that the maxmimum number of mutations possible is still five. To ensure a fair comparison, we ensured both feature steering and simulated annealing took a comparable amount of time. We set the number of simulated annealing timesteps based on the time required for feature steering to design 50 variants. This was done by first measuring the time needed to complete 1,000 simulated annealing timesteps and then scaling accordingly.

\textbf{Random.} To create the random baseline, we sample from the Poisson distribution of $\text{Pois}(2) + 1$ to determine the number of mutations to make. We then choose the mutated amino acid uniformly at random. 

\subsection{MLP Training}
\label{appendix:mlp}

To create a fitness prediction model for our protein engineering tasks, we trained an MLP for each DMS assay. The MLP is a three-layer feedforward network with ReLU activation functions, taking in a flattened one-hot encoding of the entire protein sequence as input. The network architecture consists of an input layer, a hidden layer with 128 neurons, a second hidden layer with 64 neurons, and a final output layer with a single neuron to predict the fitness score.

For each DMS assay, we split the full data into a training set (80\%) and a validation set (20\%). We then trained the MLP for up to 1000 epochs using the AdamW optimizer with a learning rate of $10^{-3}$ and Mean Squared Error (MSE) as the loss function. To prevent overfitting, we employed early stopping with a patience of 10 epochs based on the validation loss. 

\subsection{Feature Visualization}
\label{appendix:feat-vis}
Given the wildtype sequence, we find the latent representation $\mathbf{z} \in \mathbb{R}^{d_{\text{SAE}} \times L}$ by passing the sequence through layer 24 of ESM2 to get the embeddings and then passing the embeddings through the SAE encoder. We use the linear probe weights and find the indices that correspond to the five largest positive and negative probe weight indices for which the corresponding index in $\mathbf{z}$ is active as well. We then find the amino acids in the sequence that are being activated by the SAE: given the $i^{\text{th}}$ latent, the amino acid activations associated with this latent are $\mathbf{z}[i,:]$. We then use the top five mutants with the highest fitness found from steering the SAE and analyze the activation difference to find the amino acids in the sequence that had the largest absolute activation difference between the wild-type and steered sequence SAE embeddings. We use PyMOL to visualize these changes.

To identify active sites in GFP, we utilize the positions provided in~\cite{weinstein2023designed} under the Methods section titled ``Refinement and mutational scan''. For GB1, we identify allosteric and binding sites based on~\cite{faure2022mapping} from Extended Data Fig. 7c. We additionally identify epistatic sites based on~\cite{olson2014comprehensive}.

\subsection{Weight Sparsity}
\label{appendix:weight-sparsity}
To quantify sparsity in linear probe weights, we measure the proportion of total variance explained by the top 5\% of weights ranked by magnitude. Using linear probe weights from the random extrapolation task with the first seed, we compute, for each training size $N$, the ratio between the variance captured by the top 5\% of weights and the total variance of all weights (where the total number of weights is $d_{\text{SAE}}$ for SAE, $d_{\text{ESM}}$ for ESM layer, and $V$ for ESM logits) is computed. We plot the magnitude of probe weights for each model in Fig.~\ref{fig:histogram}. For visualization purposes, we exclude weights that have a magnitude greater than 3. This occurs 11 times in the ESM logits but not in the ESM layer or SAE.

\section{Additional Experimental Results}
\label{appendix:expt_results}

\begin{table*}[h]
\vspace{0.2cm}
\caption{Average Spearman $\rho$ across all low-$N$ regimes under mutation extrapolation.}
\label{tab:mutation}
\centering
\resizebox{1 \textwidth}{!}{%
\begin{tabular}{lrcccc}
\toprule
\textbf{Method} & \textbf{DMS}  & \textbf{$N$ = 8 $\uparrow$}  & \textbf{$N$ = 24 $\uparrow$}  & \textbf{$N$ = 96 $\uparrow$}  & \textbf{$N$ = 384 $\uparrow$} \\
\midrule
\multirow{6}{*}{SAE}  & GFP\_AEQVI\_Sarkisyan & 0.06 $\pm$ 0.09 & \textbf{0.15 $\pm$ 0.10} & \textbf{0.29 $\pm$ 0.08} & \textbf{0.36 $\pm$ 0.04} \\
  & SPG1\_STRSG\_Olson & 0.15 $\pm$ 0.13 & \textbf{0.42 $\pm$ 0.09} & \textbf{0.67 $\pm$ 0.05} & \textbf{0.79 $\pm$ 0.02} \\
  & SPG1\_STRSG\_Wu & 0.14 $\pm$ 0.25 & \textbf{0.31 $\pm$ 0.21} & \textbf{0.34 $\pm$ 0.11} & \textbf{0.46 $\pm$ 0.13} \\
  & DLG4\_HUMAN\_Faure & \textbf{0.32 $\pm$ 0.13} & \textbf{0.39 $\pm$ 0.07} & \textbf{0.50 $\pm$ 0.09} & \textbf{0.61 $\pm$ 0.05} \\
  & GRB2\_HUMAN\_Faure & \textbf{0.33 $\pm$ 0.22} & \textbf{0.49 $\pm$ 0.06} & \textbf{0.63 $\pm$ 0.04} & \textbf{0.67 $\pm$ 0.03} \\
  & F7YBW8\_MESOW\_Ding & \textbf{0.54 $\pm$ 0.23} & \textbf{0.57 $\pm$ 0.16} & 0.61 $\pm$ 0.17 & 0.63 $\pm$ 0.20 \\
\cmidrule(lr){1-6}
\multirow{6}{*}{ESM layer}  & GFP\_AEQVI\_Sarkisyan & 0.05 $\pm$ 0.09 & 0.11 $\pm$ 0.11 & 0.23 $\pm$ 0.06 & 0.29 $\pm$ 0.07 \\
  & SPG1\_STRSG\_Olson & \textbf{0.20 $\pm$ 0.19} & 0.34 $\pm$ 0.07 & 0.63 $\pm$ 0.04 & 0.78 $\pm$ 0.02 \\
  & SPG1\_STRSG\_Wu & \textbf{0.17 $\pm$ 0.32} & 0.30 $\pm$ 0.23 & 0.30 $\pm$ 0.09 & 0.33 $\pm$ 0.22 \\
  & DLG4\_HUMAN\_Faure & 0.22 $\pm$ 0.14 & 0.33 $\pm$ 0.08 & 0.47 $\pm$ 0.07 & 0.55 $\pm$ 0.11 \\
  & GRB2\_HUMAN\_Faure & \textbf{0.33 $\pm$ 0.22} & 0.44 $\pm$ 0.08 & 0.61 $\pm$ 0.04 & \textbf{0.67 $\pm$ 0.03} \\
  & F7YBW8\_MESOW\_Ding & 0.51 $\pm$ 0.23 & 0.54 $\pm$ 0.21 & 0.59 $\pm$ 0.17 & \textbf{0.64 $\pm$ 0.20} \\
\cmidrule(lr){1-6}
\multirow{6}{*}{ESM logits}  & GFP\_AEQVI\_Sarkisyan & \textbf{0.13 $\pm$ 0.10} & 0.13 $\pm$ 0.13 & 0.22 $\pm$ 0.06 & 0.30 $\pm$ 0.03 \\
  & SPG1\_STRSG\_Olson & 0.16 $\pm$ 0.12 & 0.29 $\pm$ 0.09 & 0.42 $\pm$ 0.06 & 0.58 $\pm$ 0.03 \\
  & SPG1\_STRSG\_Wu & 0.10 $\pm$ 0.13 & 0.07 $\pm$ 0.24 & 0.08 $\pm$ 0.23 & 0.26 $\pm$ 0.28 \\
  & DLG4\_HUMAN\_Faure & 0.13 $\pm$ 0.11 & 0.15 $\pm$ 0.08 & 0.22 $\pm$ 0.12 & 0.34 $\pm$ 0.10 \\
  & GRB2\_HUMAN\_Faure & 0.15 $\pm$ 0.17 & 0.30 $\pm$ 0.07 & 0.50 $\pm$ 0.08 & 0.59 $\pm$ 0.03 \\
  & F7YBW8\_MESOW\_Ding & 0.37 $\pm$ 0.41 & 0.52 $\pm$ 0.32 & \textbf{0.64 $\pm$ 0.19} & \textbf{0.64 $\pm$ 0.20} \\
\cmidrule(lr){1-6}
\bottomrule
\end{tabular}%
}
\end{table*}

\begin{table*}[h]
\vspace{0.2cm}
\caption{Average Spearman $\rho$ across all low-$N$ regimes under position extrapolation.}
\label{tab:position}
\centering
\resizebox{1 \textwidth}{!}{%
\begin{tabular}{lrcccc}
\toprule
\textbf{Method} & \textbf{DMS}  & \textbf{$N$ = 8 $\uparrow$}  & \textbf{$N$ = 24 $\uparrow$}  & \textbf{$N$ = 96 $\uparrow$}  & \textbf{$N$ = 384 $\uparrow$} \\
\midrule
\multirow{6}{*}{SAE}  & GFP\_AEQVI\_Sarkisyan & 0.10 $\pm$ 0.09 & 0.13 $\pm$ 0.11 & \textbf{0.25 $\pm$ 0.09} & 0.26 $\pm$ 0.15 \\
  & SPG1\_STRSG\_Olson & \textbf{0.18 $\pm$ 0.15} & 0.36 $\pm$ 0.28 & \textbf{0.54 $\pm$ 0.10} & \textbf{0.65 $\pm$ 0.09} \\
  & SPG1\_STRSG\_Wu & 0.11 $\pm$ 0.32 & 0.08 $\pm$ 0.26 & \textbf{0.15 $\pm$ 0.28} & \textbf{0.25 $\pm$ 0.23} \\
  & DLG4\_HUMAN\_Faure & \textbf{0.37 $\pm$ 0.20} & \textbf{0.52 $\pm$ 0.10} & \textbf{0.53 $\pm$ 0.10} & 0.57 $\pm$ 0.08 \\
  & GRB2\_HUMAN\_Faure & \textbf{0.28 $\pm$ 0.20} & 0.27 $\pm$ 0.24 & \textbf{0.51 $\pm$ 0.08} & 0.48 $\pm$ 0.11 \\
  & F7YBW8\_MESOW\_Ding & \textbf{0.09 $\pm$ 0.46} & 0.10 $\pm$ 0.47 & 0.06 $\pm$ 0.50 & 0.29 $\pm$ 0.33 \\
\cmidrule(lr){1-6}
\multirow{6}{*}{ESM layer}  & GFP\_AEQVI\_Sarkisyan & 0.04 $\pm$ 0.10 & 0.09 $\pm$ 0.09 & 0.23 $\pm$ 0.06 & 0.25 $\pm$ 0.12 \\
  & SPG1\_STRSG\_Olson & 0.18 $\pm$ 0.19 & \textbf{0.42 $\pm$ 0.32} & 0.46 $\pm$ 0.18 & 0.55 $\pm$ 0.15 \\
  & SPG1\_STRSG\_Wu & \textbf{0.16 $\pm$ 0.29} & \textbf{0.12 $\pm$ 0.24} & 0.14 $\pm$ 0.40 & 0.20 $\pm$ 0.30 \\
  & DLG4\_HUMAN\_Faure & 0.15 $\pm$ 0.38 & 0.41 $\pm$ 0.24 & 0.41 $\pm$ 0.23 & \textbf{0.58 $\pm$ 0.07} \\
  & GRB2\_HUMAN\_Faure & 0.25 $\pm$ 0.21 & 0.24 $\pm$ 0.21 & 0.40 $\pm$ 0.21 & 0.40 $\pm$ 0.09 \\
  & F7YBW8\_MESOW\_Ding & 0.05 $\pm$ 0.54 & 0.07 $\pm$ 0.51 & 0.25 $\pm$ 0.37 & 0.05 $\pm$ 0.38 \\
\cmidrule(lr){1-6}
\multirow{6}{*}{ESM logits}  & GFP\_AEQVI\_Sarkisyan & \textbf{0.12 $\pm$ 0.05} & \textbf{0.13 $\pm$ 0.10} & 0.21 $\pm$ 0.09 & \textbf{0.26 $\pm$ 0.07} \\
  & SPG1\_STRSG\_Olson & 0.17 $\pm$ 0.20 & 0.30 $\pm$ 0.19 & 0.41 $\pm$ 0.16 & 0.50 $\pm$ 0.16 \\
  & SPG1\_STRSG\_Wu & 0.13 $\pm$ 0.26 & 0.03 $\pm$ 0.19 & 0.14 $\pm$ 0.27 & 0.05 $\pm$ 0.16 \\
  & DLG4\_HUMAN\_Faure & 0.00 $\pm$ 0.14 & 0.14 $\pm$ 0.15 & 0.26 $\pm$ 0.20 & 0.39 $\pm$ 0.10 \\
  & GRB2\_HUMAN\_Faure & 0.20 $\pm$ 0.16 & \textbf{0.27 $\pm$ 0.14} & 0.41 $\pm$ 0.10 & \textbf{0.49 $\pm$ 0.08} \\
  & F7YBW8\_MESOW\_Ding & 0.07 $\pm$ 0.51 & \textbf{0.44 $\pm$ 0.33} & \textbf{0.34 $\pm$ 0.38} & \textbf{0.39 $\pm$ 0.19} \\
\cmidrule(lr){1-6}
\bottomrule
\end{tabular}%
}
\end{table*}

\begin{table*}[h]
\vspace{0.2cm}
\caption{Average Spearman $\rho$ across all low-$N$ regimes under regime extrapolation.}
\label{tab:regime}
\centering
\resizebox{1 \textwidth}{!}{%
\begin{tabular}{lrcccc}
\toprule
\textbf{Method} & \textbf{DMS}  & \textbf{$N$ = 8 $\uparrow$}  & \textbf{$N$ = 24 $\uparrow$}  & \textbf{$N$ = 96 $\uparrow$}  & \textbf{$N$ = 384 $\uparrow$} \\
\midrule
\multirow{6}{*}{SAE}  & GFP\_AEQVI\_Sarkisyan & \textbf{0.11 $\pm$ 0.10} & \textbf{0.23 $\pm$ 0.13} & \textbf{0.36 $\pm$ 0.05} & \textbf{0.56 $\pm$ 0.05} \\
  & SPG1\_STRSG\_Olson & 0.21 $\pm$ 0.14 & \textbf{0.39 $\pm$ 0.06} & \textbf{0.69 $\pm$ 0.05} & \textbf{0.84 $\pm$ 0.01} \\
  & SPG1\_STRSG\_Wu & \textbf{0.14 $\pm$ 0.16} & 0.22 $\pm$ 0.11 & \textbf{0.27 $\pm$ 0.10} & \textbf{0.32 $\pm$ 0.04} \\
  & DLG4\_HUMAN\_Faure & \textbf{0.31 $\pm$ 0.20} & \textbf{0.39 $\pm$ 0.14} & \textbf{0.58 $\pm$ 0.09} & 0.67 $\pm$ 0.06 \\
  & GRB2\_HUMAN\_Faure & \textbf{0.34 $\pm$ 0.11} & 0.39 $\pm$ 0.15 & 0.65 $\pm$ 0.07 & \textbf{0.77 $\pm$ 0.01} \\
  & F7YBW8\_MESOW\_Ding & \textbf{0.37 $\pm$ 0.20} & 0.53 $\pm$ 0.08 & 0.60 $\pm$ 0.06 & 0.65 $\pm$ 0.03 \\
\cmidrule(lr){1-6}
\multirow{6}{*}{ESM layer}  & GFP\_AEQVI\_Sarkisyan & 0.06 $\pm$ 0.10 & 0.17 $\pm$ 0.15 & 0.32 $\pm$ 0.06 & 0.49 $\pm$ 0.04 \\
  & SPG1\_STRSG\_Olson & \textbf{0.23 $\pm$ 0.09} & 0.38 $\pm$ 0.08 & 0.64 $\pm$ 0.05 & 0.83 $\pm$ 0.01 \\
  & SPG1\_STRSG\_Wu & 0.14 $\pm$ 0.18 & \textbf{0.23 $\pm$ 0.11} & 0.22 $\pm$ 0.08 & 0.29 $\pm$ 0.06 \\
  & DLG4\_HUMAN\_Faure & 0.25 $\pm$ 0.14 & 0.34 $\pm$ 0.17 & 0.49 $\pm$ 0.13 & \textbf{0.70 $\pm$ 0.05} \\
  & GRB2\_HUMAN\_Faure & 0.30 $\pm$ 0.09 & \textbf{0.40 $\pm$ 0.14} & \textbf{0.65 $\pm$ 0.06} & 0.76 $\pm$ 0.01 \\
  & F7YBW8\_MESOW\_Ding & 0.35 $\pm$ 0.17 & 0.52 $\pm$ 0.09 & 0.59 $\pm$ 0.05 & \textbf{0.68 $\pm$ 0.02} \\
\cmidrule(lr){1-6}
\multirow{6}{*}{ESM logits}  & GFP\_AEQVI\_Sarkisyan & 0.11 $\pm$ 0.27 & 0.17 $\pm$ 0.15 & 0.24 $\pm$ 0.17 & 0.40 $\pm$ 0.04 \\
  & SPG1\_STRSG\_Olson & 0.16 $\pm$ 0.13 & 0.25 $\pm$ 0.09 & 0.42 $\pm$ 0.03 & 0.56 $\pm$ 0.02 \\
  & SPG1\_STRSG\_Wu & 0.04 $\pm$ 0.09 & 0.14 $\pm$ 0.07 & 0.16 $\pm$ 0.05 & 0.23 $\pm$ 0.03 \\
  & DLG4\_HUMAN\_Faure & 0.16 $\pm$ 0.12 & 0.27 $\pm$ 0.07 & 0.34 $\pm$ 0.09 & 0.36 $\pm$ 0.07 \\
  & GRB2\_HUMAN\_Faure & 0.05 $\pm$ 0.12 & 0.32 $\pm$ 0.06 & 0.48 $\pm$ 0.04 & 0.59 $\pm$ 0.03 \\
  & F7YBW8\_MESOW\_Ding & 0.35 $\pm$ 0.14 & \textbf{0.56 $\pm$ 0.08} & \textbf{0.60 $\pm$ 0.04} & 0.64 $\pm$ 0.02 \\
\cmidrule(lr){1-6}
\bottomrule
\end{tabular}%
}
\end{table*}

\begin{table*}[h]
\vspace{0.2cm}
\caption{Average Spearman $\rho$ across all low-$N$ regimes under score extrapolation.}
\label{tab:score}
\centering
\resizebox{1 \textwidth}{!}{%
\begin{tabular}{lrcccc}
\toprule
\textbf{Method} & \textbf{DMS}  & \textbf{$N$ = 8 $\uparrow$}  & \textbf{$N$ = 24 $\uparrow$}  & \textbf{$N$ = 96 $\uparrow$}  & \textbf{$N$ = 384 $\uparrow$} \\
\midrule
\multirow{6}{*}{SAE}  & GFP\_AEQVI\_Sarkisyan & 0.01 $\pm$ 0.09 & 0.01 $\pm$ 0.05 & 0.02 $\pm$ 0.04 & \textbf{0.02 $\pm$ 0.03} \\
  & SPG1\_STRSG\_Olson & 0.04 $\pm$ 0.13 & 0.03 $\pm$ 0.10 & 0.12 $\pm$ 0.10 & 0.09 $\pm$ 0.04 \\
  & SPG1\_STRSG\_Wu & 0.07 $\pm$ 0.07 & \textbf{0.04 $\pm$ 0.07} & 0.10 $\pm$ 0.09 & 0.18 $\pm$ 0.05 \\
  & DLG4\_HUMAN\_Faure & \textbf{0.05 $\pm$ 0.09} & 0.04 $\pm$ 0.08 & 0.01 $\pm$ 0.06 & 0.06 $\pm$ 0.05 \\
  & GRB2\_HUMAN\_Faure & 0.01 $\pm$ 0.06 & \textbf{0.16 $\pm$ 0.11} & \textbf{0.20 $\pm$ 0.07} & 0.27 $\pm$ 0.05 \\
  & F7YBW8\_MESOW\_Ding & \textbf{0.17 $\pm$ 0.24} & 0.00 $\pm$ 0.31 & 0.22 $\pm$ 0.20 & 0.27 $\pm$ 0.10 \\
\cmidrule(lr){1-6}
\multirow{6}{*}{ESM layer}  & GFP\_AEQVI\_Sarkisyan & \textbf{0.01 $\pm$ 0.06} & 0.01 $\pm$ 0.05 & 0.01 $\pm$ 0.04 & 0.01 $\pm$ 0.04 \\
  & SPG1\_STRSG\_Olson & \textbf{0.06 $\pm$ 0.08} & \textbf{0.03 $\pm$ 0.08} & \textbf{0.12 $\pm$ 0.08} & \textbf{0.13 $\pm$ 0.05} \\
  & SPG1\_STRSG\_Wu & \textbf{0.11 $\pm$ 0.11} & 0.02 $\pm$ 0.08 & \textbf{0.11 $\pm$ 0.08} & \textbf{0.23 $\pm$ 0.06} \\
  & DLG4\_HUMAN\_Faure & 0.01 $\pm$ 0.09 & \textbf{0.04 $\pm$ 0.06} & \textbf{0.03 $\pm$ 0.07} & \textbf{0.07 $\pm$ 0.07} \\
  & GRB2\_HUMAN\_Faure & \textbf{0.02 $\pm$ 0.07} & 0.14 $\pm$ 0.11 & 0.18 $\pm$ 0.06 & \textbf{0.30 $\pm$ 0.04} \\
  & F7YBW8\_MESOW\_Ding & 0.14 $\pm$ 0.24 & 0.03 $\pm$ 0.31 & 0.23 $\pm$ 0.19 & 0.37 $\pm$ 0.09 \\
\cmidrule(lr){1-6}
\multirow{6}{*}{ESM logits}  & GFP\_AEQVI\_Sarkisyan & 0.01 $\pm$ 0.08 & \textbf{0.05 $\pm$ 0.08} & \textbf{0.06 $\pm$ 0.04} & 0.00 $\pm$ 0.06 \\
  & SPG1\_STRSG\_Olson & 0.00 $\pm$ 0.11 & 0.01 $\pm$ 0.09 & 0.09 $\pm$ 0.10 & 0.05 $\pm$ 0.06 \\
  & SPG1\_STRSG\_Wu & 0.02 $\pm$ 0.06 & 0.03 $\pm$ 0.07 & 0.08 $\pm$ 0.05 & 0.13 $\pm$ 0.05 \\
  & DLG4\_HUMAN\_Faure & 0.05 $\pm$ 0.10 & 0.01 $\pm$ 0.07 & 0.02 $\pm$ 0.04 & 0.00 $\pm$ 0.04 \\
  & GRB2\_HUMAN\_Faure & 0.02 $\pm$ 0.08 & 0.04 $\pm$ 0.08 & 0.12 $\pm$ 0.05 & 0.23 $\pm$ 0.04 \\
  & F7YBW8\_MESOW\_Ding & 0.15 $\pm$ 0.30 & \textbf{0.03 $\pm$ 0.22} & \textbf{0.30 $\pm$ 0.21} & \textbf{0.39 $\pm$ 0.14} \\
\cmidrule(lr){1-6}
\bottomrule
\end{tabular}%
}
\end{table*}

\end{document}